\documentclass[letterpaper, 10 pt, conference]{ieeeconf}  

\IEEEoverridecommandlockouts                              

\usepackage{graphics} 
\usepackage{epsfig} 
\usepackage{amsmath} 
\usepackage{amssymb}  
\usepackage[table,xcdraw]{xcolor}
 \usepackage{booktabs}
\usepackage{multirow}
\usepackage{algorithm}
\usepackage{algorithmic}
\definecolor{agg_g}{RGB}{112,173,71}
\definecolor{agg_r}{RGB}{255,0,0}
\usepackage[noadjust]{cite}
\usepackage[font={footnotesize}]{caption}
\usepackage{hyperref}       

\graphicspath{{images/}}
\usepackage{hhline}

\newcommand{\ie}{\textit{i}.\textit{e}. }

\title{Diving Deeper Into Pedestrian Behavior Understanding: Intention Estimation, Action Prediction, and Event Risk Assessment}
\author{Amir Rasouli$^{1}$ and Iuliia Kotseruba$^{2}$
\thanks{$^{1}$Huawei Technologies Canada, {\tt amir.rasouli@huawei.com}}%
\thanks{$^{2}$York University, {\tt\ yulia84@yorku.ca}. Work was done while at Huawei Technologies Canada.}}%

\begin{document}

\maketitle

\begin{abstract}
In this paper, we delve into the pedestrian behavior understanding problem from the perspective of three different tasks: intention estimation, action prediction, and event risk assessment. We first define the tasks and discuss how these tasks are represented and annotated in two widely used pedestrian datasets, JAAD and PIE. We then propose a new benchmark based on these definitions, available annotations, and three new classes of metrics, each designed to assess different aspects of the model performance. 

We apply the new evaluation approach to examine four SOTA prediction models on each task and compare their performance w.r.t. metrics and input modalities. In particular, we analyze the differences between intention estimation and action prediction tasks by considering various scenarios and contextual factors. Lastly, we examine model agreement across these two tasks to show their complementary role. The proposed benchmark reveals new facts about the role of different data modalities, the tasks, and relevant data properties. We conclude by elaborating on our findings and proposing future research directions\footnote{Code is available at \href{https://github.com/aras62/PIE/tree/master/scenarioEval}{github.com/aras62/PIE/tree/master/scenarioEval}}.
\end{abstract}

\section{Introduction}
Safety is the primary concern for predicting pedestrian behavior in traffic. The problem can be formulated as determining whether the pedestrian's action will lead them to appear in the path of the vehicle. There is a growing number of solutions to this problem that aim to anticipate pedestrians' actions (e.g., crossing the road) from monocular videos and vehicle sensors. Although benchmark datasets established for this task continue to register robust performance improvements, there remain some outstanding issues. 

One of the ongoing concerns is the conflation of intention and action prediction tasks in the literature. Particularly, after the introduction of datasets that provide data for both tasks (e.g., \cite{rasouli2019pie, chen2021psi}), it has become difficult to discern models trained for intention estimation and action prediction as the terms are often used interchangeably. Additionally, these tasks only indicate potential risk, but on their own are not sufficient to measure the direct impact of the predicted events on the behavior of the intelligent vehicle.

Another issue is the narrow focus of evaluation procedures that measure performance by averaging accuracy of models over all observations. For safety purposes, a deeper understanding of the algorithm performance is needed, particularly, because most models are difficult to interpret. For example, it is important to assess how early the models can forecast future actions and how consistent the predictions remain as the vehicle approaches the pedestrian. 

\begin{figure}
    \centering
    \includegraphics[width=1\columnwidth]{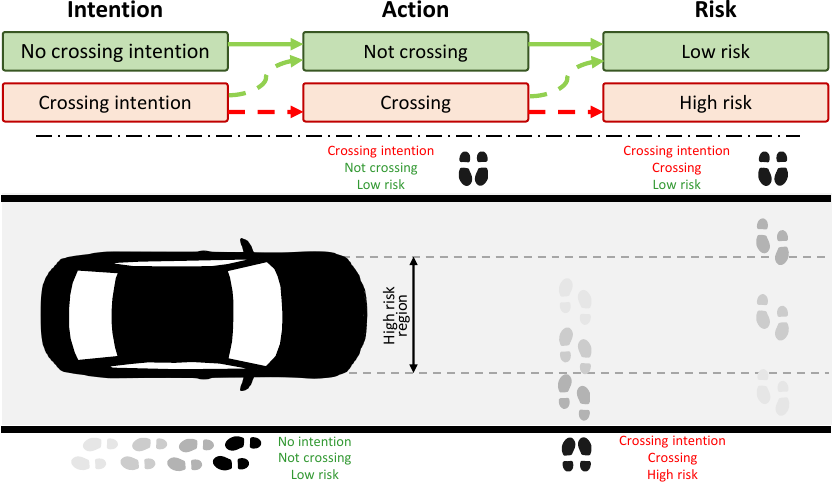}
    \caption{An overview of different tasks of pedestrian behavior understanding. \textbf{Top:} connections between different tasks---definite (solid arrows) and probable (dashed arrows). \textbf{Bottom:} examples of pedestrians with different types of behavior and associated risks.}\vspace{-0.5cm}
    \label{fig:first-fig}
\end{figure}

The contributions of this paper are summarized as follows: 1) We provide a formal definition for intention and action prediction tasks; 2) We introduce an event risk assessment task designed to measure the impact of the predicted action on the ego-vehicle (see Figure \ref{fig:first-fig}); 3) We propose new metrics that focus on measuring how timely, balanced, and consistent are model predictions; 4) We evaluate state-of-the-art (SOTA) models on all three tasks, with particular focus on highlighting the differences between intention estimation and action prediction, identifying what factors impact each task, and assessing model agreement on both. 

\section{Related Work}
\subsection{Task definitions}
\label{sec:task_definitions}

We define the following three tasks related to understanding pedestrian behavior in traffic: \textit{intention estimation}, \textit{action prediction}, and \textit{event risk assessment}. They follow in this order: first, the pedestrian decides to cross (intention), begins crossing the road if circumstances permit (action), which may or may not put them in a way of the ego-vehicle (risk) as shown in Figure \ref{fig:first-fig}. 

\noindent
\textbf{Intention vs. action.} The difference between forming a goal and acting on it was already established in the 1890s \cite{james1890principles} and became a part of several theories of human behavior \cite{searle1983intentionality, bratman1987intention}, as well as more recent implementations \cite{georgeff1999belief, schneemann2016context, rasouli2019pie}. Following these works, we consider \textit{crossing intention} a precursor of action. Intention is a state of mind, so it cannot be observed directly, but may lead to action under certain conditions. In contrast, \textit{crossing action} is an observable event of the pedestrian crossing the road in front of the ego-vehicle.  

In the literature, ``intent(ion) prediction'' \cite{saleh2019real, gujjar2019classifying,  liu2020spatiotemporal, sui2021joint, yang2021crossing, wu2021applying, chen2021visual, razali2021pedestrian, zhang2021pedestrian, naik2022scene, song2022pedestrian,  yang2022predicting, huang2023learning, zhou2023pit, upreti2023traffic} and ``action prediction'' \cite{rasouli2017they, rasouli2020pedestrian, cadena2019pedestrian, chaabane2020looking, yau2021graph, yao2021coupling, rasouli2021bifold, singh2021multi, gesnouin2022assessing, achaji2022attention, rasouli2022multi, cadena2022pedestrian, zhai2022social, zhang2023cross, rasouli2023pedformer} occur with almost equal frequency but, with few exceptions \cite{sui2021joint,zhai2022social,zhang2023cross}, both terms mean predicting pedestrian \textit{actions}. Here, we refer to the tasks as \textit{intention estimation} since intention exists in the present, and \textit{action prediction}, as it is concerned with future events. In intelligent driving, intention (e.g. in the form of the goal or target of the agent \cite{karim2023destine,Wang_2023_ICRA_1}) is often used in conjunction with action prediction for improved accuracy. Here, we evaluate these tasks separately to highlight their differences and identify the challenges pertaining to each task.

\noindent
\textbf{Event risk assessment.} Assessing the risk posed by other vehicles or pedestrians is crucial for intelligent driving systems \cite{xiao2023review,wang2022potential,herman2021pedestrian}. Pedestrian intention and upcoming action can indicate the possibility of risky events, but on their own they are not sufficient to measure the impact of those events on the intelligent vehicle. Trajectory prediction models directly estimate future locations of pedestrians, as a sequence of coordinates \cite{rasouli2023novel,rasouli2019pie} or/and final destination \cite{rasouli2023pedformer,rasouli2021bifold}, but further interpretation is needed to determine their potential risk. For example, one can estimate whether the pedestrian will end up in the driver’s comfort zone \cite{herman2021pedestrian}. Here, we extend this idea to the egocentric setting and propose to directly assess the future risk of pedestrian action with respect to the ego-vehicle based on the risk regions in the image plane that are aligned with the center of the ego-vehicle.

\subsection{Model evaluation}
A number of datasets for studying and modeling pedestrian behavior have been proposed \cite{rasouli2017they, rasouli2019pie, malla2020titan,liu2020spatiotemporal, rasouli2020pepscenes, chen2021psi, guo2022pedestrian}, out of which JAAD \cite{rasouli2017they} and PIE \cite{rasouli2019pie} are currently the most used. The majority of models trained and evaluated on these datasets \cite{zhou2023pit,upreti2023traffic,yang2022predicting,song2022pedestrian} follow the protocol in \cite{kotseruba2021benchmark}.

JAAD and PIE provide multi-modal data consisting of monocular video footage filmed from inside the moving vehicle and annotations: spatial (bounding boxes for pedestrians and relevant objects, pedestrian poses), textual (labels describing properties of the scene, pedestrian behaviors and attributes, and drivers' actions), and numeric (vehicle telemetry). Existing models rely on a variety of input modalities, including visual context \cite{rasouli2017they, saleh2019real, gujjar2019classifying, chaabane2020looking, liu2020spatiotemporal}, pedestrian poses \cite{singh2021multi}, or bounding boxes \cite{achaji2022attention}, or a combination of these modalities for more robust performance \cite{rasouli2022multi, yang2022predicting, kotseruba2021benchmark}. In this work, we evaluate models with different input modalities to highlight performance differences on the proposed tasks.

Past evaluation approaches relied on a subset of classification metrics, such as accuracy, recall, precision, AUC, and F1-score. In addition, results of individual models were related to various aspects of the data, such as time-to-event \cite{rasouli2020pedestrian, kotseruba2021benchmark}, observation length \cite{rasouli2020pedestrian, cadena2022pedestrian}, prediction horizon \cite{liu2020spatiotemporal, achaji2022attention}, input features \cite{rasouli2020pedestrian, naik2022scene}, scale and occlusion of bounding boxes \cite{kotseruba2021benchmark}, and ego-vehicle speed \cite{upreti2023traffic}. However, in all cases, metrics are averaged over all samples across different time horizons and pedestrian instances. Such evaluation assesses the overall performance, but fails to address consistency of the models and their limitations in predicting different horizons or risk levels. Here, we propose several additional metrics to capture the latter aspects of the models.

\section{Experiment Setup}
We evaluate four SOTA action prediction models, SFGRU \cite{rasouli2020pedestrian}, BiPed \cite{rasouli2021bifold}, PCPA \cite{kotseruba2021benchmark}, and PedFormer \cite{rasouli2023pedformer}, on two public benchmarks -- Pedestrian Intention Estimation (PIE) dataset \cite{rasouli2019pie} and JAAD \cite{rasouli2017they}.  Below, we discuss data processing, model properties, and metrics definitions.

\subsection{Data}
\noindent
\textbf{Action and intention annotations.} Both datasets contain annotated videos of traffic scenes.  Two types of annotations are the most relevant here: pedestrian \textit{intentions} (which reflect their motivation to cross) and \textit{crossing actions} (that specify whether they will cross in front of the ego-vehicle). 

Because intentions are not directly observable, in PIE, intention labels were aggregated from the responses of human subjects who viewed the clips from the dataset and indicated whether pedestrians in them \textit{intended} (or wanted) to cross the street (not necessarily in front of the ego-vehicle). These scores were averaged and used as intention labels. Note that these intention labels are not ground truth per se, but rather a probabilistic estimation of pedestrians' intentions. 

Intent labels in JAAD are binary (not probabilistic) and are assigned as follows: non-crossing intent is assigned to all bystander pedestrians, \ie those that do not interact with the ego-vehicle or are deemed irrelevant by the annotators, and the rest are considered as having a crossing intent. Due to the obvious biases in the labeling process and lack of experimental validation (as in PIE), these labels do not effectively reflect intentions of pedestrians and therefore will not be used for the experiments.

Crossing actions in both datasets simply state whether the pedestrian was observed crossing the road in front of the ego-vehicle, however, intention annotations are different. 

\vspace{1em}
\noindent
\textbf{Data split.} For all tasks, we set the observation length to $0.5s$ ($15$ frames) and extract samples with a $30 \%$ overlap to get a more uniform distribution. Note that for the remainder of the paper, \textit{instance} refers to the entire track of the individual pedestrian and \textit{sample} refers to a portion of this track, comprised of \textit{observation} and \textit{prediction}.

\begin{figure*}
    \centering
    \includegraphics[width=1\textwidth]{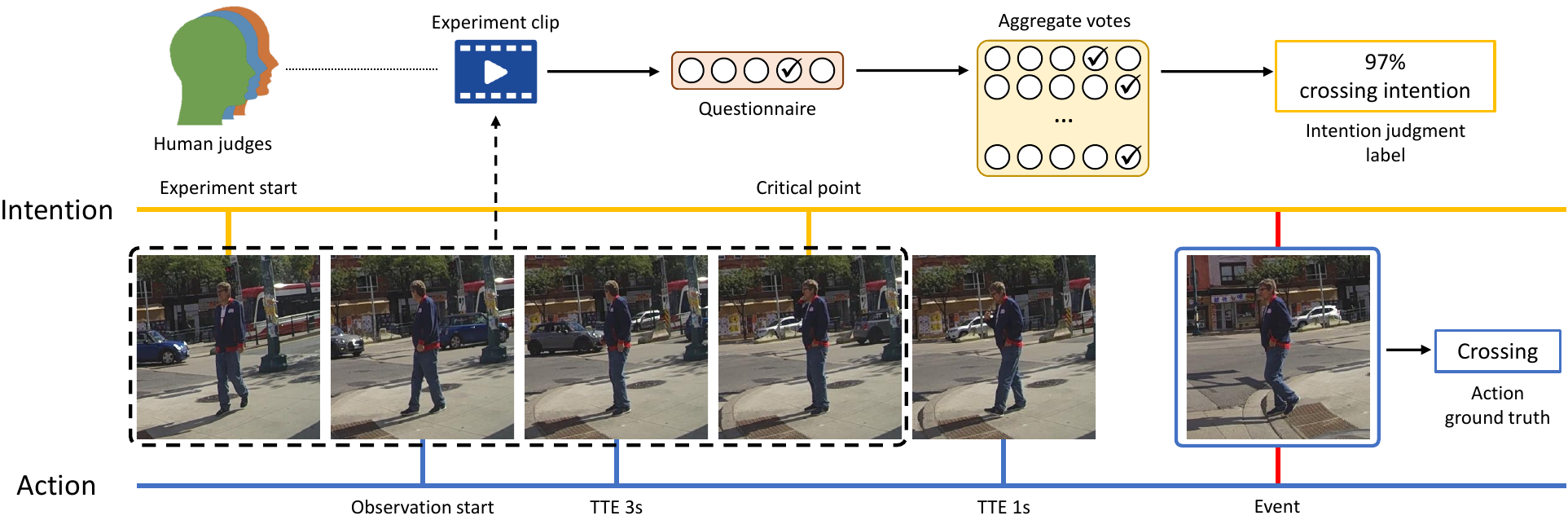}    \vspace{-1.5em}
    \caption{Overview of annotations and sampling in PIE. Intention labels are represented by aggregated votes of human observers who watched videos of pedestrians from \textit{experiment start} up to the \textit{critical point}. Action labels are based on the observed action of crossing in front of the ego-vehicle. Sequences for action prediction task are sampled so that the observations end between $1$-$3s$ TTE. \textit{Observation start} is the earliest frame that is fed to the model.}
    \label{fig:int_act_annot}
    \vspace{-1.5em}
\end{figure*}
\label{sec:data_split}

\subsubsection{Intention estimation} As mentioned earlier, in PIE, intention estimation labels are collected from human subjects who viewed short clips of pedestrians extracted from the dataset. The start and end points of these clips are specified in the annotations with \texttt{exp\_start\_point} and \texttt{critical\_point} tags for each pedestrian instance (see Figure \ref{fig:int_act_annot}). We use these points to sample data and discard instances shorter than observation length.  

Given the probabilistic nature of intention labels ($\mathrm{intention} \in [0,1]$), we divide them into three equal intervals for no-crossing intention (NCI), unsure intention (UI) and crossing intention (CI), respectively (see Table \ref{tbl:data_stats}).

\begin{table}[]
\centering
\caption{Number of samples for intention and action tasks.}\label{tbl:data_stats}    \vspace{-0.5em}
 \resizebox{\columnwidth}{!}{\begin{tabular}{
>{\columncolor[HTML]{FFFFFF}}c 
>{\columncolor[HTML]{FFFFFF}}c 
>{\columncolor[HTML]{FFFFFF}}c 
>{\columncolor[HTML]{FFFFFF}}c 
>{\columncolor[HTML]{FFFFFF}}c 
>{\columncolor[HTML]{FFF2CC}}c 
>{\columncolor[HTML]{FFFFFF}}l 
>{\columncolor[HTML]{FFFFFF}}c 
>{\columncolor[HTML]{FFFFFF}}c 
>{\columncolor[HTML]{FFFFFF}}c 
>{\columncolor[HTML]{FFFFFF}}c 
>{\columncolor[HTML]{FFFFFF}}c 
>{\columncolor[HTML]{FFF2CC}}c }
\multicolumn{6}{c}{\cellcolor[HTML]{FFFFFF}\textbf{Intention}}                                                                                                                                                                                                                                                                                      &  & \multicolumn{6}{c}{\cellcolor[HTML]{FFFFFF}\textbf{Action}}                                                                                                                                                                                                \\
\hhline{~~|*{4}{-}~~~*{4}{-}|} 
\multicolumn{1}{l}{\cellcolor[HTML]{FFFFFF}}                                & \multicolumn{1}{l}{\cellcolor[HTML]{FFFFFF}}                & \textbf{Train}                               & \textbf{Test}                                & \multicolumn{1}{c|}{\cellcolor[HTML]{FFFFFF}\textbf{Val}}  & \textbf{\# Ped}                                       &  & \multicolumn{1}{l}{\cellcolor[HTML]{FFFFFF}}                                 &                                                             & \textbf{Train} & \textbf{Test} & \multicolumn{1}{c|}{\cellcolor[HTML]{FFFFFF}\textbf{Val}}  & \textbf{\# Ped} \\ \hhline{-|*{5}{-}~-*{5}{-}|} 
\multicolumn{1}{c|}{\cellcolor[HTML]{FFFFFF}}                               & \multicolumn{1}{c|}{\cellcolor[HTML]{FFFFFF}\textbf{NCI}}   & 1922                                         & 756                                          & \multicolumn{1}{c|}{\cellcolor[HTML]{FFFFFF}449}           & 329                                          &  & \multicolumn{1}{c|}{\cellcolor[HTML]{FFFFFF}}                                & \multicolumn{1}{c|}{\cellcolor[HTML]{FFFFFF}\textbf{NC}}    & 4163           & 3203          & \multicolumn{1}{c|}{\cellcolor[HTML]{FFFFFF}1218}          & 1282            \\
\multicolumn{1}{c|}{\cellcolor[HTML]{FFFFFF}}                               & \multicolumn{1}{c|}{\cellcolor[HTML]{FFFFFF}\textbf{UI}}    & 1009                                         & 835                                          & \multicolumn{1}{c|}{\cellcolor[HTML]{FFFFFF}312}           & 220                                          &  & \multicolumn{1}{c|}{\cellcolor[HTML]{FFFFFF}}                                & \multicolumn{1}{c|}{\cellcolor[HTML]{FFFFFF}\textbf{C}}     & 1417           & 1255          & \multicolumn{1}{c|}{\cellcolor[HTML]{FFFFFF}327}           & 499             \\ 
\hhline{~|~|~~~|
        >{\arrayrulecolor[HTML]{FFF2CC}}-
        >{\arrayrulecolor[HTML]{000000}}~~|*{5}{-}|} 
\multicolumn{1}{c|}{\cellcolor[HTML]{FFFFFF}}                               & \multicolumn{1}{c|}{\cellcolor[HTML]{FFFFFF}\textbf{CI}}    & 5282                                         & 4881                                         & \multicolumn{1}{c|}{\cellcolor[HTML]{FFFFFF}1538}          & 1285                                       &  & \multicolumn{1}{c|}{\multirow{-3}{*}{\cellcolor[HTML]{FFFFFF}\rotatebox{90}{\textbf{PIE}}}}  & \multicolumn{1}{c|}{\cellcolor[HTML]{FFFFFF}\textbf{Total}} & \textbf{5580}  & \textbf{4458} & \multicolumn{1}{c|}{\cellcolor[HTML]{FFFFFF}\textbf{1545}} & \textbf{1781}   \\ \hhline{~|*{5}{-}|~-|*{5}{-}|} 
\multicolumn{1}{c|}{\multirow{-4}{*}{\cellcolor[HTML]{FFFFFF}\rotatebox{90}{\textbf{PIE}}}} & \multicolumn{1}{c|}{\cellcolor[HTML]{FFFFFF}\textbf{Total}} & \textbf{8213}                                & \textbf{6472}                                & \multicolumn{1}{c|}{\cellcolor[HTML]{FFFFFF}\textbf{2299}} & \textbf{1834}                                &  & \multicolumn{1}{c|}{\cellcolor[HTML]{FFFFFF}}                                & \multicolumn{1}{c|}{\cellcolor[HTML]{FFFFFF}\textbf{NC}}    & 4456           & 3548          & \multicolumn{1}{c|}{\cellcolor[HTML]{FFFFFF}702}           & 1519            \\
\multicolumn{1}{l}{\cellcolor[HTML]{FFFFFF}}                                & \multicolumn{1}{l}{\cellcolor[HTML]{FFFFFF}}                & \multicolumn{1}{l}{\cellcolor[HTML]{FFFFFF}} & \multicolumn{1}{l}{\cellcolor[HTML]{FFFFFF}} & \multicolumn{1}{l}{\cellcolor[HTML]{FFFFFF}}               & \multicolumn{1}{l}{\cellcolor[HTML]{FFFFFF}} &  & \multicolumn{1}{c|}{\cellcolor[HTML]{FFFFFF}}                                & \multicolumn{1}{c|}{\cellcolor[HTML]{FFFFFF}\textbf{C}}     & 1122           & 769           & \multicolumn{1}{c|}{\cellcolor[HTML]{FFFFFF}107}           & 353             \\ 
\hhline{*{8}{~}|*{5}{-}|} 
\multicolumn{1}{l}{\cellcolor[HTML]{FFFFFF}}                                & \multicolumn{1}{l}{\cellcolor[HTML]{FFFFFF}}                & \multicolumn{1}{l}{\cellcolor[HTML]{FFFFFF}} & \multicolumn{1}{l}{\cellcolor[HTML]{FFFFFF}} & \multicolumn{1}{l}{\cellcolor[HTML]{FFFFFF}}               & \multicolumn{1}{l}{\cellcolor[HTML]{FFFFFF}} &  & \multicolumn{1}{c|}{\multirow{-3}{*}{\cellcolor[HTML]{FFFFFF}\rotatebox{90}{\textbf{JAAD}}}} & \multicolumn{1}{c|}{\textbf{Total}}                                              & \textbf{5578}  & \textbf{4317} & \multicolumn{1}{c|}{\cellcolor[HTML]{FFFFFF}\textbf{809}}  & \textbf{1872}  
\end{tabular}}\vspace{-0.7cm}
\end{table}
\subsubsection{Action prediction} Crossing events in both PIE and JAAD datasets are labeled as \texttt{crossing\_point}. This tag indicates the frame when the pedestrian started crossing or the last frame the pedestrian was visible if they did not cross in front of the ego-vehicle. We extract samples with time-to-event (TTE) between $1$ to $3s$ ($30$ to $90$ frames) (see Figure \ref{fig:int_act_annot}) and only exclude samples below TTE of $1s$ and instances shorter than observation length (see Table \ref{tbl:data_stats}).

\subsubsection{Event risk assessment} We divide the image plane into equal vertical regions $160$ pixels wide (double the average width of the pedestrians' bounding boxes). As a result, there are a total of $12$ regions representing 6 classes of risk (due to symmetry, as shown in Figure \ref{fig:risk_class}). The prediction horizon is set to $3s$, double the average reaction time to surprise events \cite{green2000long}. The risk class is assigned based on the center coordinate of the bounding box at the end of the prediction horizon, calculated from the last observation frame. If for a given pedestrian the bounding box coordinate at prediction time does not exist (i.e. the pedestrian is not visible anymore), the last available bounding box is selected.
\vspace{-0.5em}
\subsection{Models}
We select models with different architectures and input modalities to highlight the differences between different design choices. BiPed \cite{rasouli2021bifold} and PedFormer \cite{rasouli2023pedformer} are multitask models that simultaneously predict trajectories and actions of pedestrians. Although architecturally different, both models rely mainly on observed trajectories, ego-vehicle sensors, and some visual context (in the form of semantic maps for interaction modeling). The other two models, SFGRU \cite{rasouli2020pedestrian} and PCPA \cite{kotseruba2021benchmark}, are single-task, i.e., predict only the probability of crossing. Besides ego-dynamics and pedestrian trajectories, they rely on visual information (actual images of pedestrians and their surroundings) and pedestrian poses.

To adapt these models for intention estimation and event risk assessment tasks, we modify their objective functions, while using default parameters. For BiPed and PedFormer, we change only the crossing action task and keep the auxiliary trajectory and grid prediction tasks the same.  
\vspace{-0.5em}
\subsection{Base metrics}
Following \cite{kotseruba2021benchmark}, we report the results using common classification metrics: accuracy ($\text{Acc}$), Area Under the Curve ($\text{AUC}$), $\text{F1}$, and precision ($\text{Prec}$). To mitigate effects of class imbalances in the datasets, we also compute balanced accuracy ($\text{bAcc}$) and mean average precision ($\text{mAP}$).
\vspace{-0.5em}
\subsection{Weighted metrics} 
\subsubsection{Action prediction} In general, prediction is easier closer to the event, as more contextual cues become available. However, in safety-critical applications like driving, it is vital to make accurate predictions as early as possible. As a result, we propose a per-sample weighted average of metrics based on the TTE of the samples. i.e., the closer a sample is to the event, the lower is its weight. We express weights using a exponential function as follows:
\begin{gather*}
       \omega_{a} = \frac{\exp^{-\frac{1}{2}\times(\frac{d_{tte}}{\sigma})^2}}{\sum_{TTE}\omega}, \text{    }
    d_{tte} = \frac{\max(TTE) - tte_a}{\max(TTE)} ,
\end{gather*}
where $\omega_a$ and $tte_a$ are the weight and TTE of the sample and $\max(TTE)$ is the maximum of TTEs across all samples, in this case $3s$. We set $\sigma = 0.3$ empirically.
\begin{figure}
    \centering
    \includegraphics[width=0.7\columnwidth]{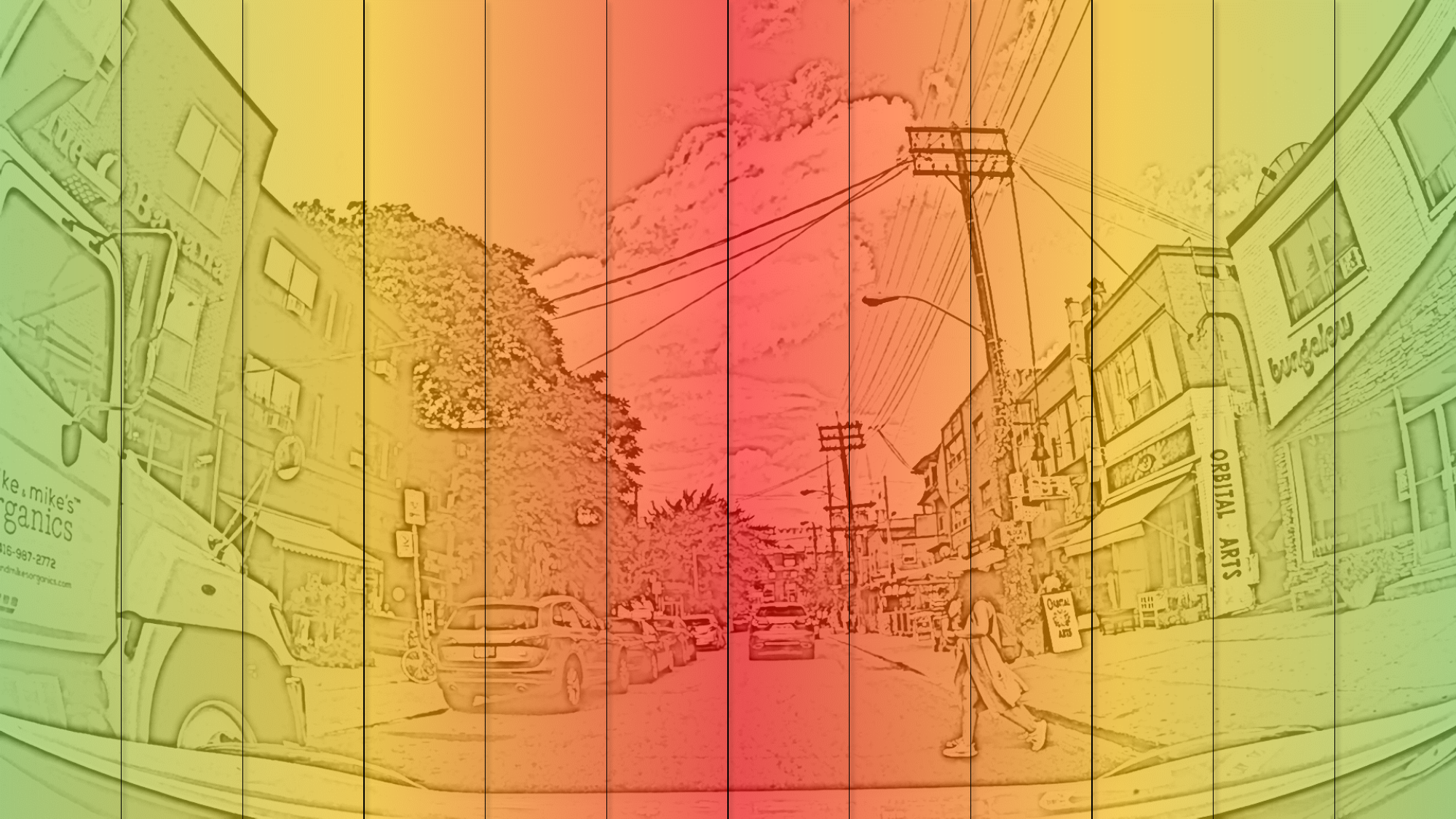}
    \caption{Example of event risk regions overlaid on the view from the ego-vehicle. Colors from red to green represent the associated risk from highest to lowest, respectively.}
    \label{fig:risk_class}
    \vspace{-1.5em}
\end{figure}

\subsubsection{Event risk assessment} Future locations of pedestrians have different implications for the ego-vehicle. Pedestrians who are directly in front of the vehicle pose more risk because they are more difficult to avoid. With this insight, we assign more weight to high risk samples ending in the center of the camera view, and gradually lower the weight towards the edges (Figure \ref{fig:risk_class}). The weights are given by, 
\begin{gather*}
       \omega_r = \exp^{-\frac{1}{2}\times(\frac{d_{cls}}{m\sigma})^2},\\
d_{cls}= 
\begin{cases}
    |cls_r - \lceil m\rceil|, & \parbox{3.2cm}{$\text{if } N_{rc} \mod 2 = 1  \lor (N_{rc} \mod 2 = 0 \land cls_r <= m$} \\
    |cls_r - \lceil m \rceil - 1|,              & \text{otherwise}
\end{cases}
\end{gather*}
where $\omega_r$ and $cls_r$ are weight and class index of the  sample, respectively, $N_{rc}$ denotes the total number of risk regions (classes), and $m = \lceil \frac{N_{rc}}{2} \rceil$.  We set $\sigma = 0.5$ empirically.

\subsection{Per-instance metrics}

\begin{figure}[!t]
    \centering
    \includegraphics[width=1\columnwidth]{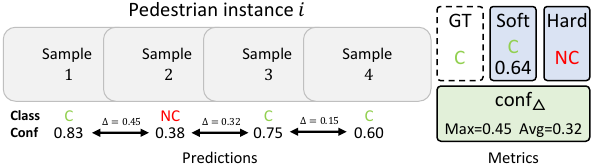}\vspace{-0.5em}
    \caption{Per-instance metric example for binary action prediction. GT refers to ground truth. Soft label is computed by averaging over prediction confidence of all samples. Hard label is set to a label other than ground truth if prediction for at least one sample does not agree with the rest, i.e. it is treated as a misprediction. If predicted labels for all samples are the same, hard label will be the same. }\vspace{-1.5em}
    \label{fig:annotations}
\end{figure}

To capture model consistency, we propose three new metrics. In order to compute them, we first rearrange the samples corresponding to each pedestrian instance in the order they have been originally extracted, i.e., resembling a moving window.  Each pedestrian instance can contain between $1$ and $N$ samples, depending on the length of the track. We then compute metrics per-instance and average them over all instances (see Figure \ref{fig:annotations}).

\subsubsection{Soft metrics} Metrics are averaged across samples corresponding to the unique pedestrian instance.

\subsubsection{Hard metrics} For each pedestrian instance, if the most confident class of all samples are the same, then that class is the prediction of that instance. Otherwise, we set the prediction of the instance as incorrect. For instance, if the correct label of a pedestrian is crossing, but the model predicted at least one of the samples as non-crossing, we set the overall prediction of that pedestrian to non-crossing.

\subsubsection{Confidence delta} We compute changes in the model's confidence score for each class between two consecutive samples and report maximum and average delta given by,
$$
\text{conf}_\Delta = \frac{\sum_{i\in \{1,...,n-1\}} |\text{conf}^{i}_c-\text{conf}^{i+1}_c|}{n-1},
$$
where $n$ is the total number of samples in that instance and $\text{conf}_c$ is the model's confidence for class $c$.

\section{Evaluation: Intention and Action}

\subsection{Performance on benchmarks}

\begin{table*}[t]
\caption{Experiment results for intention estimation in PIE.  $\uparrow$ and $\downarrow$ mean higher or lower values are better respectively.}\label{tbl:intention_results}\vspace{-0.5em}
\centering
\begin{tabular}{lcccccc|cccc|c}
                                    & \multicolumn{10}{c}{\textbf{PIE}}& \multicolumn{1}{l}{}                       \\ \cline{2-12} 
& \multicolumn{6}{c|}{\textbf{Base} $\uparrow$}  & \multicolumn{4}{c|}{\textbf{Soft/Hard} $\uparrow$} & \multicolumn{1}{l}{\textbf{Max/Avg} $\downarrow$ } \\ \cline{2-12} 
\multicolumn{1}{l|}{\textbf{Model}}& $\text{mAP}$ & $\text{bAcc}$& $\text{AUC}$&$\text{Acc}$& $\text{Prec}$ & $\text{F1} $ & $\text{Acc}$& $\text{bAcc}$ & $\text{Prec}$& $\text{F1}$& $\text{conf}_\Delta$\\ \hline
\multicolumn{1}{l|}{\textbf{SFGRU}} & 0.44&  0.41& 0.65&\textbf{0.76}& \textbf{0.52}& 0.41& \cellcolor[HTML]{DDEBF7}0.76/\textbf{0.67} & \cellcolor[HTML]{DDEBF7}0.41/\textbf{0.31} & \cellcolor[HTML]{DDEBF7}\textbf{0.78}/0.28 & \cellcolor[HTML]{DDEBF7}0.41/\textbf{0.29} & \cellcolor[HTML]{E2EFDA}\textbf{0.14}/\textbf{0.04}   \\
\multicolumn{1}{l|}{\textbf{PCPA}}& \textbf{0.46}& 0.42& 0.65& 0.75& 0.41& 0.41& \cellcolor[HTML]{DDEBF7}\textbf{0.77}/0.59 & \cellcolor[HTML]{DDEBF7}\textbf{0.45}/0.27 & \cellcolor[HTML]{DDEBF7}0.45/0.28 & \cellcolor[HTML]{DDEBF7}\textbf{0.44}/0.27 & \cellcolor[HTML]{E2EFDA}0.21/0.06    \\
\multicolumn{1}{l|}{\textbf{BiPed}}& 0.42&  0.41& 0.66&0.73& 0.40& 0.39& \cellcolor[HTML]{DDEBF7}0.75/0.57 & \cellcolor[HTML]{DDEBF7}0.40/0.27 & \cellcolor[HTML]{DDEBF7}0.38/0.29 & \cellcolor[HTML]{DDEBF7}0.38/0.27 & \cellcolor[HTML]{E2EFDA}0.30/0.09    \\
\multicolumn{1}{l|}{\textbf{PedFormer}} & \textbf{0.46}&  \textbf{0.45} & \textbf{0.70}&0.70& 0.43& \textbf{0.43}& \cellcolor[HTML]{DDEBF7}0.72/0.55 & \cellcolor[HTML]{DDEBF7}0.44/0.28 & \cellcolor[HTML]{DDEBF7}0.42/\textbf{0.31} & \cellcolor[HTML]{DDEBF7}0.42/\textbf{0.29} & \cellcolor[HTML]{E2EFDA}0.23/0.06   
\end{tabular}\vspace{-1em}
\end{table*}

\begin{table*}[t]
\caption{Experiment results for action prediction.  $\uparrow$ and $\downarrow$ mean higher or lower values are better respectively..}\label{tbl:action_results} \vspace{-0.5em}
\centering
\begin{tabular}{lccccccccccc}
                                        & \multicolumn{10}{c}{\textbf{PIE}}& \multicolumn{1}{l}{}                       \\ \cline{2-12} 
\textbf{}                               & \multicolumn{3}{c|}{\textbf{Base} $\uparrow$}                                 & \multicolumn{3}{c|}{\textbf{Base/Weighted} $\uparrow$}                                       & \multicolumn{4}{c|}{\textbf{Soft/Hard} $\uparrow$}                                                                                                                                                                                                        & \textbf{Max/Avg} $\downarrow$                           \\ \cline{2-12} 
\multicolumn{1}{l|}{\textbf{Model}}     & $\text{mAP}$  & $\text{bAcc}$ & \multicolumn{1}{c|}{$\text{AUC}$}  & $\text{Acc}$       & $\text{Prec}$      & \multicolumn{1}{c|}{$\text{F1}$}        & $\text{bAcc}$                                                  & $\text{Acc}$                                                   & $\text{Prec}$                              & \multicolumn{1}{c|}{$\text{F1}$}                                &  $\text{conf}_\Delta$                             \\ \hline
\multicolumn{1}{l|}{\textbf{SFGRU}}     & 0.75          & 0.75          & \multicolumn{1}{c|}{0.87}          & 0.83/0.82          & 0.79/0.73          & \multicolumn{1}{c|}{0.65/0.64}          & \cellcolor[HTML]{DDEBF7}0.76/0.61                              & \cellcolor[HTML]{DDEBF7}0.85/0.72                              & \cellcolor[HTML]{DDEBF7}0.87/0.50          & \multicolumn{1}{c|}{\cellcolor[HTML]{DDEBF7}0.67/0.43}          & \cellcolor[HTML]{E2EFDA}\textbf{0.10/0.04} \\
\multicolumn{1}{l|}{\textbf{PCPA}}      & 0.79          & 0.81          & \multicolumn{1}{c|}{0.89}          & 0.86/0.85          & 0.81/0.75          & \multicolumn{1}{c|}{0.74/0.72}          & \cellcolor[HTML]{DDEBF7}0.81/0.63                              & \cellcolor[HTML]{DDEBF7}0.88/0.72                              & \cellcolor[HTML]{DDEBF7}\textbf{0.90}/0.50 & \multicolumn{1}{c|}{\cellcolor[HTML]{DDEBF7}0.76/0.46}          & \cellcolor[HTML]{E2EFDA}0.17/0.07          \\
\multicolumn{1}{l|}{\textbf{BiPed}}     & 0.84          & 0.84          & \multicolumn{1}{c|}{0.93}          & \textbf{0.89/0.88} & 0.84/0.80          & \multicolumn{1}{c|}{\textbf{0.79}/0.77} & \cellcolor[HTML]{DDEBF7}\textbf{0.86}/0.66                    & \cellcolor[HTML]{DDEBF7}0.90/0.74                              & \cellcolor[HTML]{DDEBF7}\textbf{0.90}/0.56 & \multicolumn{1}{c|}{\cellcolor[HTML]{DDEBF7}\textbf{0.82}/0.51} & \cellcolor[HTML]{E2EFDA}0.16/0.06          \\
\multicolumn{1}{l|}{\textbf{PedFormer}} & \textbf{0.88} & \textbf{0.85} & \multicolumn{1}{c|}{\textbf{0.94}} & \textbf{0.89/0.88} & \textbf{0.85/0.81} & \multicolumn{1}{c|}{\textbf{0.79/0.78}} & \cellcolor[HTML]{DDEBF7}\textbf{0.86/0.76}                     & \cellcolor[HTML]{DDEBF7}\textbf{0.91/0.80}                     & \cellcolor[HTML]{DDEBF7}0.89/\textbf{0.66} & \multicolumn{1}{c|}{\cellcolor[HTML]{DDEBF7}\textbf{0.82/0.65}} & \cellcolor[HTML]{E2EFDA}0.12/\textbf{0.04} \\ \hline
                                        & \multicolumn{10}{c}{}  &   \\ 
                                        & \multicolumn{10}{c}{\textbf{JAAD}}  &   \\ \hline
\multicolumn{1}{l|}{\textbf{SFGRU}}     & 0.61          & 0.76          & \multicolumn{1}{c|}{0.85}          & \textbf{0.86/0.87} & \textbf{0.60/0.63} & \multicolumn{1}{c|}{0.61/\textbf{0.61}} & \multicolumn{1}{l}{\cellcolor[HTML]{DDEBF7}0.74/\textbf{0.60}} & \multicolumn{1}{l}{\cellcolor[HTML]{DDEBF7}0.86/\textbf{0.74}} & \cellcolor[HTML]{DDEBF7}0.62/\textbf{0.34} & \multicolumn{1}{c|}{\cellcolor[HTML]{DDEBF7}0.59/0.31}          & \cellcolor[HTML]{E2EFDA}0.17/\textbf{0.07} \\
\multicolumn{1}{l|}{\textbf{PCPA}}      & 0.56          & 0.72          & \multicolumn{1}{c|}{0.81}          & 0.83/0.83          & 0.51/0.51          & \multicolumn{1}{c|}{0.54/0.53}          & \multicolumn{1}{l}{\cellcolor[HTML]{DDEBF7}0.72/0.55}          & \multicolumn{1}{l}{\cellcolor[HTML]{DDEBF7}0.84/0.70}          & \cellcolor[HTML]{DDEBF7}0.54/0.27          & \multicolumn{1}{c|}{\cellcolor[HTML]{DDEBF7}0.54/0.24}          & \cellcolor[HTML]{E2EFDA}0.16/\textbf{0.07} \\
\multicolumn{1}{l|}{\textbf{BiPed}}     & 0.60          & 0.75          & \multicolumn{1}{c|}{0.86}          & 0.85/0.85          & 0.57/0.58          & \multicolumn{1}{c|}{0.58/0.59}          & \multicolumn{1}{l}{\cellcolor[HTML]{DDEBF7}0.75/0.52}          & \multicolumn{1}{l}{\cellcolor[HTML]{DDEBF7}\textbf{0.87}/0.68} & \cellcolor[HTML]{DDEBF7}\textbf{0.68}/0.23 & \multicolumn{1}{c|}{\cellcolor[HTML]{DDEBF7}0.61/0.20}          & \cellcolor[HTML]{E2EFDA}0.21/0.10          \\
\multicolumn{1}{l|}{\textbf{PedFormer}} & \textbf{0.63} & \textbf{0.78} & \multicolumn{1}{c|}{\textbf{0.87}} & \textbf{0.86}/0.85 & 0.58/0.58          & \multicolumn{1}{c|}{\textbf{0.62/0.61}} & \multicolumn{1}{l}{\cellcolor[HTML]{DDEBF7}\textbf{0.78}/0.58} & \multicolumn{1}{l}{\cellcolor[HTML]{DDEBF7}\textbf{0.87}/0.72} & \cellcolor[HTML]{DDEBF7}0.64/0.32          & \multicolumn{1}{c|}{\cellcolor[HTML]{DDEBF7}\textbf{0.64/0.28}} & \cellcolor[HTML]{E2EFDA}\textbf{0.15/0.07}
\end{tabular}\vspace{-1.5em}
\end{table*}

\textbf{Intention estimation} results in Tables \ref{tbl:intention_results} show that the performance of the models differ. While PedFormer stands out on most base metrics, on others it lags significantly behind SFGRU -- by up to $6\%$ in accuracy and $9\%$ in precision. Hard and soft metrics highlight other differences. For instance, PCPA and SFGRU that rely more on visual context clearly dominate. Of particular interest is $\text{soft precision}$ of SFGRU which is $33\%$ higher than the next best model, PCPA. This shows that SFGRU is fairly successful at distinguishing between pedestrian instances of different intention classes. 

Hard metrics show significant performance drop on all models, suggesting that their overall consistency is low, i.e. predicted intentions fluctuate across successive samples within the pedestrian instances. But the amount of fluctuation varies and once again SFGRU is the best with the lowest max and avg $\text{conf}_\Delta$, $7\%$ and $2\%$ better compared to  PCPA.

\textbf{Action prediction} results, shown in Table \ref{tbl:action_results}, tell a different story. On PIE, the more dynamics oriented model, PedFormer, is the best on almost all metrics, and $\text{hard}$ metrics in particular, where it performs at least $6\%$ better on \text{Acc} and up to $14\%$ better on $\text{F1}$. This indicates that pedestrian trajectories and ego-motion are important for predicting upcoming actions. On the same dataset, SFGRU remains the most consistent model, having the best $\text{conf}_\Delta$ but with the lowest hard scores. 

On  JAAD, which only has categorical ego-motion information, the results are mixed. PedFormer is still better on most base metrics, whereas SFGRU does better on $\text{Acc}$ and more so on $\text{Prec}$. While $\text{soft}$ metrics favor PedFormer, SFGRU is more successful in $\text{hard}$ metrics, showing better instance-wise consistency. On $\text{conf}_\Delta$, all models perform very similarly, except for BiPed, which is more inconsistent.

The difference between base metrics and their weighted counterparts for most models is marginal, with some exceptions on JAAD, where weighted metrics are better. This can be due to noise or general inconsistencies in JAAD annotations that in some cases favor samples extracted further away from the time of event points. On PIE, weighted precision of some models is lower than base by up to $6\%$. This is expected, as the uncertainty of prediction is higher farther away from the crossing event.

\subsection{Impact of context modeling on performance} Given the high variability of traffic scenes and black-box nature of deep learning models, it is generally difficult to pinpoint what contextual elements contributed to the correct predictions. However, considering the architectural differences between the models and their performance on different tasks and datasets, we can observe some patterns. 

Referring back to the task definitions in Section \ref{sec:task_definitions}, pedestrian intention reflects their motivation or goal, which is not affected by the environmental factors. For instance, if a pedestrian \textit{wants} to cross the road to go to a store, they will try to do so either at the signalized crossing or by finding a safe gap in traffic if the nearest controlled intersection is too far. However, the pedestrian's ultimate intention (or goal) to cross the road remains constant, unless their objective of going to the store on the other side of the street changes.

In terms of modeling context, different algorithms rely on different sources of information. While SFGRU  and PCPA use images of pedestrians (cropped to capture surrounding context) and their poses, BiPed and PedFormer mainly rely on dynamics (pedestrians' and the ego-vehicle's) and only use visual context represented by semantic maps of the scenes for modeling interactions between the agents.

On the intention task, performance of all models is generally low due to the inherent difficulty of the task and the uncertainty and noise present in human judgment annotations. However, models that rely more on visual features tend to be more successful at distinguishing different intention classes. Thus, in addition to dynamic cues, it is likely that detailed visual context, e.g. head orientation, posture, etc., is necessarily for accurate estimation.

In comparison, results on action prediction show that effective modeling of scene dynamics is more important for prediction accuracy. This is apparent in the ranking of the models in Tables \ref{tbl:intention_results} and \ref{tbl:action_results} on the PIE dataset. On JAAD which lacks accurate dynamics information, we can see  a significant degradation of the performance on action prediction on all metrics.  

\subsection{Scenario-based analysis}

\begin{table}[]
\centering
\caption{The $\text{mAP}$ for intention estimation and action prediction tasks on PIE for different scenarios. The colors are computed over all cells for each task. Green and red indicate the best and worst performance, respectively.}\label{tbl:scen_1_results}\vspace{-0.7em}
\includegraphics[width=0.9\columnwidth]{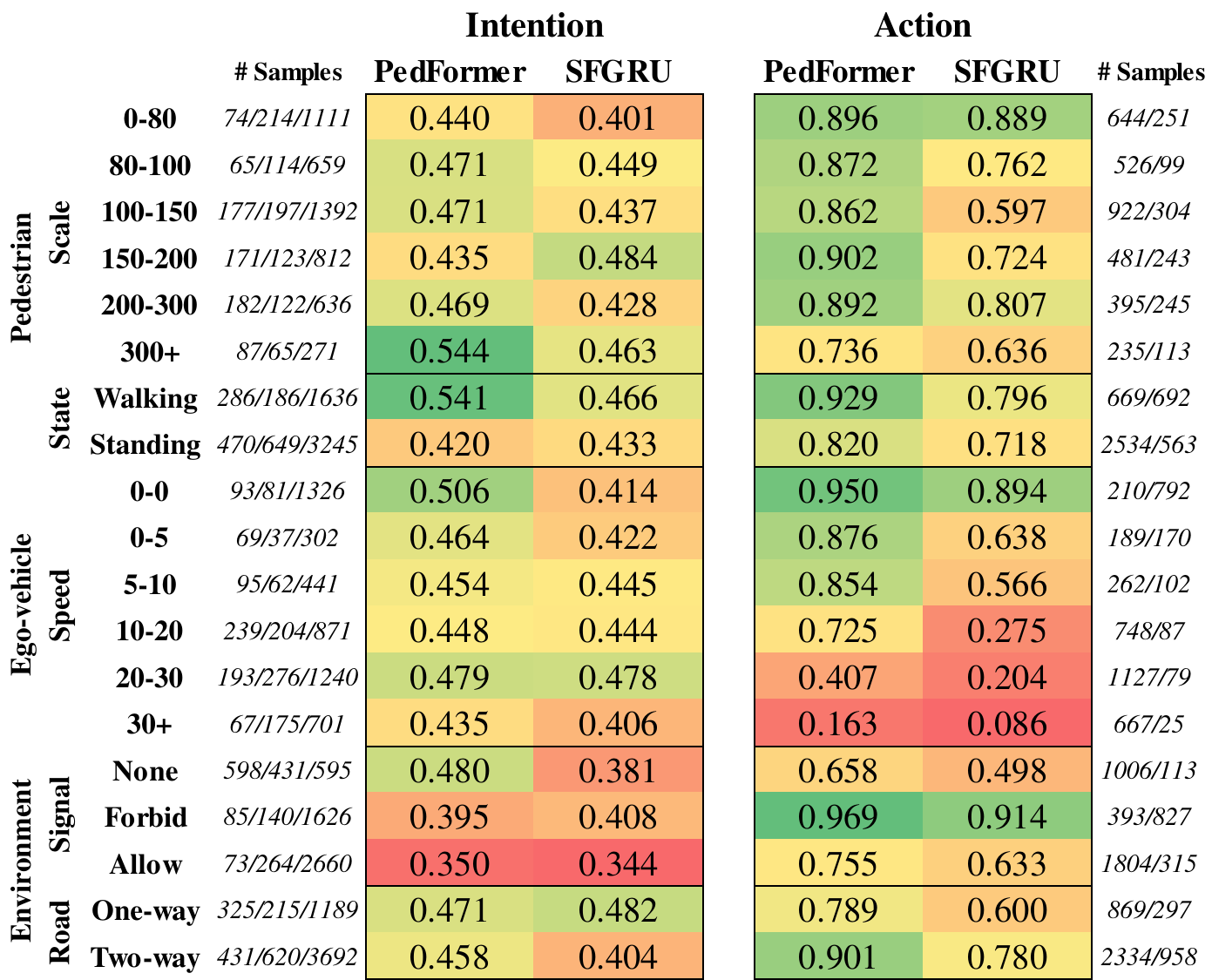}\vspace{-1.5em}
\end{table}

To further highlight the differences between intention estimation and action prediction tasks, we take a closer look at the data properties. We use the PIE dataset and based on its annotations, split scenarios into three categories, \textbf{pedestrian}, \textbf{ego-vehicle}, and \textbf{environment}. For pedestrians, we consider two factors: \textit{scale} equal to the height of bounding boxes in pixels and \textit{state} indicating whether the pedestrian is walking or standing during the observation period. In the case of the ego-vehicle, we consider \textit{speed} in km/h. For environment, we use \textit{signal} and \textit{road}. Signal indicates the traffic light state w.r.t. the ego-vehicle: forbid (red), allow (yellow or green), or none (no traffic light is present). Road specifies the direction of traffic: one-way or two-way. 

For all factors, we average the characteristics over the observation period. We then select the best performing models for each task and calculate their $\text{mAP}$. The results in Table \ref{tbl:scen_1_results} show distinct impacts of different contextual factors on intention estimation and action prediction as color distributions are reflecting model performance on each task.

\textbf{Pedestrian factors.}
Pedestrian state is the most notable factor that plays an important role for both tasks.  In particular, walking towards the road is a strong indicator of crossing intention as well as likely occurrence of crossing event (consistent with the finding that walking pedestrians tend to accept shorter gaps \cite{Rasouli_2019_ITS}). The intention and upcoming action of standing pedestrians are more difficult to estimate, thus the significant drop in the model performance. 

There are some differences observable on pedestrian scale factor. For instance, PedFormer achieves the best performance on the largest scale on intention and worst on action. As there is no clear pattern of change across different scales, such performance difference can be due to the presence of other factors or perhaps the differences in the distribution of samples between intention and action tasks. 

\textbf{Ego-vehicle factors.} On the action prediction task, there is a very clear pattern of performance degradation across different ego-speed thresholds: from a high of $95\%$ on scenarios where the ego-vehicle is stationary to a low of $8\%$ when it is moving fast. This can be attributed to the impact of ego-motion on how pedestrian movements appear in the image plane, as well as increased uncertainty in pedestrian decision-making as they need to negotiate with the ego-vehicle in order to cross. On the intention task, however, the changes due to ego-speed are much smaller, only a few percent. This supports our earlier claim that behaviors of other road users should have a minor, if any, impact on the intentions of pedestrians. 

\textbf{Environment factors.} Traffic light forbid (red) state has a significant impact on the action prediction accuracy: when the ego-vehicle is not moving, its influence on pedestrians' behavior is minimized. The action prediction models also perform better on two-way streets. This can potentially be due to the properties of the dataset, in which the percentage of challenging scenarios, e.g., jaywalking, is smaller compared to narrower one-way streets. In comparison, on the intention task, there is no significant performance gap across different environment factors, pointing to the fact that intention is less influenced by them.

It should be noted that some fluctuations within scenarios and across tasks can be due to the uneven distribution of samples and the fact that not all samples have overlaps (as shown in Figure \ref{fig:int_act_annot}). In addition, a single factor analysis may not reveal all dependencies because factors may interact. For example, in ego-speed scenarios different data partitions may include pedestrians of different scales or with different states. However, a multi-factors analysis (similar to \cite{rasouli2023novel}) was not feasible due to the sparsity of the data in each subclass of the tasks for training and evaluation. 

\subsection{Model agreement between intention and action}
\begin{table}
    \centering
        \caption{Intention estimation and action prediction agreement. The columns in the shaded area correspond to joint results for intention (I) and action(A) and the colors indicate whether predictions were \textcolor{agg_g}{correct} or \textcolor{agg_r}{incorrect}. For instance, \textcolor{agg_g}{I}-\textcolor{agg_r}{A} means only intention was predicted correctly. }\vspace{-1em}
    \label{tbl:agreement}
    \includegraphics[width=0.55\columnwidth]{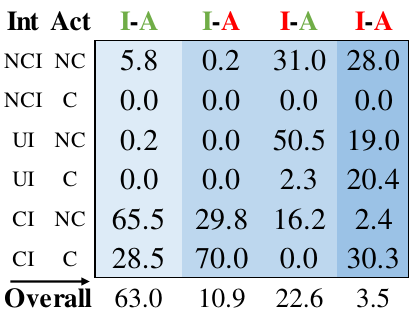}\vspace{-1.5em}
\end{table}

In this section, we test whether models' predictions for both tasks are in agreement, e.g., if the model predicts that the pedestrian has no intention of crossing, will it also predict that they will not cross the road? To test this, we train PedFormer on both intention and action tasks simultaneously. For consistency with human judgment annotations, we use the samples from the intention task (see Figure \ref{fig:int_act_annot}). 

As shown in Table \ref{tbl:agreement}, we split the results based on the combination of intention and action classes (rows) and PedFormer's performance (correct or incorrect) on each task (columns). Overall, the model infers both tasks correctly on $63\%$ of the samples, both incorrectly on $3.5\%$, and only one of the tasks correctly on the remainder.  

In $11\%$ of the partially agreeing samples, intention but not action is predicted correctly, meaning that some cues that help estimate intention do not necessarily reflect whether action will take place.  Therefore, in such cases, intention can play a complementary role. 

\begin{table*}[th!]
\caption{The results for the event risk assessment task.  $\uparrow$ and $\downarrow$ mean higher or lower values are better, respectively.}
\label{tbl:risk_results}\vspace{-0.6em}
\begin{tabular}{lccccccccccc}
                                        & \multicolumn{11}{c}{\textbf{PIE}}                                                                                                                                                                                                                                                                                                                                                                            \\ \cline{2-12} 
                                        & \multicolumn{3}{c|}{\textbf{Base} $\uparrow$}                                 & \multicolumn{3}{c|}{\textbf{Base/Weighted} $\uparrow$}                                       & \multicolumn{4}{c|}{\textbf{Soft/Hard} $\uparrow$}                                                                                                                                                                & \textbf{Max/Avg} $\downarrow$                           \\ \cline{2-12} 
\multicolumn{1}{l|}{\textbf{Model}}     & $\text{mAP}$  & $\text{bAcc}$ & \multicolumn{1}{c|}{$\text{AUC}$}  & $\text{Acc}$       & $\text{Prec}$      & \multicolumn{1}{c|}{$\text{F1}$}        & $\text{bAcc}$                               & $\text{Acc}$                              & $\text{Prec}$                              & \multicolumn{1}{c|}{$\text{F1}$}                                &  $\text{conf}_\Delta$                             \\ \hline
\multicolumn{1}{l|}{\textbf{SFGRU}}     & 0.26          & 0.22          & \multicolumn{1}{c|}{0.86}          & 0.72/0.50          & 0.23/0.21          & \multicolumn{1}{c|}{0.21/0.19}          & \cellcolor[HTML]{DDEBF7}0.23/0.16          & \cellcolor[HTML]{DDEBF7}0.75/0.64          & \cellcolor[HTML]{DDEBF7}0.25/0.16          & \multicolumn{1}{c|}{\cellcolor[HTML]{DDEBF7}0.23/0.14}          & \cellcolor[HTML]{E2EFDA}0.18/0.02          \\
\multicolumn{1}{l|}{\textbf{PCPA}}      & 0.24          & 0.18          & \multicolumn{1}{c|}{0.86}          & 0.70/0.47          & 0.17/0.16          & \multicolumn{1}{c|}{0.16/0.13}          & \cellcolor[HTML]{DDEBF7}0.17/0.15          & \cellcolor[HTML]{DDEBF7}0.72/0.62          & \cellcolor[HTML]{DDEBF7}0.15/0.15          & \multicolumn{1}{c|}{\cellcolor[HTML]{DDEBF7}0.16/0.14}          & \cellcolor[HTML]{E2EFDA}0.20/0.02          \\
\multicolumn{1}{l|}{\textbf{BiPed}}     & 0.26          & 0.23          & \multicolumn{1}{c|}{0.87}          & 0.73/0.51          & 0.28/0.25          & \multicolumn{1}{c|}{0.24/0.21}          & \cellcolor[HTML]{DDEBF7}0.21/0.15          & \cellcolor[HTML]{DDEBF7}0.75/0.65          & \cellcolor[HTML]{DDEBF7}0.22/0.20          & \multicolumn{1}{c|}{\cellcolor[HTML]{DDEBF7}0.20/0.14}          & \cellcolor[HTML]{E2EFDA}0.23/0.02          \\
\multicolumn{1}{l|}{\textbf{PedFormer}} & \textbf{0.45} & \textbf{0.42} & \multicolumn{1}{c|}{\textbf{0.95}} & \textbf{0.80/0.65} & \textbf{0.49/0.49} & \multicolumn{1}{c|}{\textbf{0.44/0.44}} & \cellcolor[HTML]{DDEBF7}\textbf{0.39/0.25} & \cellcolor[HTML]{DDEBF7}\textbf{0.82/0.69} & \cellcolor[HTML]{DDEBF7}\textbf{0.45/0.44} & \multicolumn{1}{c|}{\cellcolor[HTML]{DDEBF7}\textbf{0.40/0.29}} & \cellcolor[HTML]{E2EFDA}\textbf{0.17/0.01} \\ \hline
\textbf{}                               & \multicolumn{11}{c}{} \\
\textbf{}                               & \multicolumn{11}{c}{\textbf{JAAD}} \\ \hline
\multicolumn{1}{l|}{\textbf{SFGRU}}     & 0.23          & 0.24          & \multicolumn{1}{c|}{0.75}          & 0.42/0.27          & 0.24/0.21          & \multicolumn{1}{c|}{0.22/0.19}          & \cellcolor[HTML]{DDEBF7}0.25/0.15          & \cellcolor[HTML]{DDEBF7}0.47/0.33          & \cellcolor[HTML]{DDEBF7}0.25/0.30          & \multicolumn{1}{c|}{\cellcolor[HTML]{DDEBF7}0.23/0.14}          & \cellcolor[HTML]{E2EFDA}0.20/0.02          \\
\multicolumn{1}{l|}{\textbf{PCPA}}      & 0.16          & 0.16          & \multicolumn{1}{c|}{0.67}          & 0.37/0.18          & 0.15/0.12          & \multicolumn{1}{c|}{0.11/0.07}          & \cellcolor[HTML]{DDEBF7}0.16/0.14          & \cellcolor[HTML]{DDEBF7}0.41/0.34          & \cellcolor[HTML]{DDEBF7}0.09/0.08          & \multicolumn{1}{c|}{\cellcolor[HTML]{DDEBF7}0.11/0.09}          & \cellcolor[HTML]{E2EFDA}\textbf{0.15/0.01} \\
\multicolumn{1}{l|}{\textbf{BiPed}}     & 0.21          & 0.22          & \multicolumn{1}{c|}{0.73}          & 0.39/0.24          & 0.21/0.18          & \multicolumn{1}{c|}{0.20/0.17}          & \cellcolor[HTML]{DDEBF7}0.21/0.12          & \cellcolor[HTML]{DDEBF7}0.43/0.29          & \cellcolor[HTML]{DDEBF7}0.19/0.20          & \multicolumn{1}{c|}{\cellcolor[HTML]{DDEBF7}0.18/0.10}          & \cellcolor[HTML]{E2EFDA}0.29/0.04          \\
\multicolumn{1}{l|}{\textbf{PedFormer}} & \textbf{0.42} & \textbf{0.40} & \multicolumn{1}{c|}{\textbf{0.90}} & \textbf{0.53/0.42} & \textbf{0.43/0.41} & \multicolumn{1}{c|}{\textbf{0.41/0.39}} & \cellcolor[HTML]{DDEBF7}\textbf{0.43/0.22} & \cellcolor[HTML]{DDEBF7}\textbf{0.60/0.36} & \cellcolor[HTML]{DDEBF7}\textbf{0.48/0.43} & \multicolumn{1}{c|}{\cellcolor[HTML]{DDEBF7}\textbf{0.44/0.25}} & \cellcolor[HTML]{E2EFDA}0.22/0.02         
\end{tabular}\vspace{-2em}
\end{table*}

In the cases where only action is correct ($22.6\%$), the samples are distributed more evenly across all three classes of intention, but for the majority no crossing action takes place. This can primarily be attributed to lack of pedestrian dynamics cues, which comprise about $80\%$ of the samples. Inferring crossing intention of standing pedestrians is more challenging and requires analysis of other contextual cues, such as proximity to the road, transit station nearby, pose, etc. In fact, $78\%$ of all unsure intention samples in the test set contain pedestrians that are mainly standing for the duration of observation. Taking all the results into account, PedFormer rarely assigns unsure labels.  Note that some variations across different categories can be due to the insufficient samples, e.g., those with non-crossing intention and crossing action.


\section{Event Risk Assessment}

As discussed in Sec. \ref{sec:data_split}, event risk assessment task determines whether a given pedestrian would pose a risk to the ego-vehicle based on how close the predicted location of the pedestrian is w.r.t. the center of the vehicle. The results are summarized in Table \ref{tbl:risk_results}. Given the nature of the task and its dependency on accurate dynamics estimation, PedFormer achieves significantly better performance on all metrics, except $\text{conf}_\Delta$ on JAAD on which PCPA stands out. 

\begin{figure}[t]
    \centering
    \includegraphics[width=1\columnwidth]{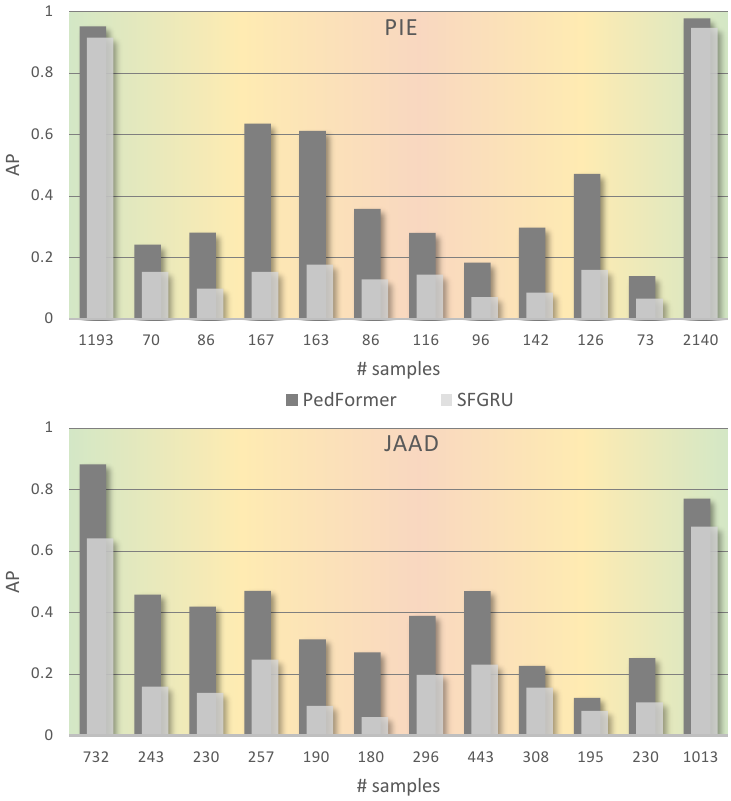}\vspace{-0.5em}
    \caption{Per-class average precision of models for the event risk assessment task. The background color in each graph represents risk associated with each region.}
    \label{fig:risk_graph}\vspace{-1.5em}
\end{figure}

Plots of per-class distribution of PedFormer and SFGRU models (see Figure \ref{fig:risk_graph}) provide a better insight into challenging areas. As anticipated, performance of the models is the best closer to the edges of the frame, partly due to more data and also the fact that pedestrians that appear there often do not cross and remain stationary at the time of prediction. In such cases, observed movements of the pedestrians in the image plane are only due to the ego-motion of the vehicle, hence the uncertainty of risk prediction is lower.

The performance on other risk classes is mixed, which can be attributed to the variability of sample properties in each class, since the number of samples in each class are about the same. Of note is the increasing gap between two models towards the center of the frame, or areas where pedestrians are crossing. Therefore, better estimation of motion is crucial for accurate prediction, hence PedFormer is more successful.

On PIE, the trends in the model performance diverge. For instance, from left to right, 4th to 5th and 6th to 7th columns (classes) the performance of PedFormer declines whereas SFGRU's improves. At the same time, PedFormer's performance drops from 5th to 6th class by more than $20\%$ and for SFGRU the difference is less than $10\%$. This indicates that besides motion information, in some scenarios, visual context is also important for accurate prediction. 

On the JAAD dataset, performance of both models trends similarly, although PedFormer is better on all classes. This is due to the fact that JAAD does not provide accurate ego-motion. As a result, both models rely mainly on the changes of pedestrian bounding boxes for reasoning about dynamics. 

\section{Discussion and Conclusions}
\subsection{Role of different tasks}
We discussed three tasks, intention estimation, action prediction, and event risk assessment, and argued that each plays a unique role in understanding and forecasting pedestrian behavior. Intention estimation reflects what an observed pedestrian wants to do. Knowing the intention, one can expect certain types of actions to follow or determine relevance of the agent, e.g., for causal representation learning \cite{Causal_agents}. Action is the realization of the intention (motive) of the pedestrian and can be seen as an early cue for possible types of motions to expect. For instance, predicted crossing action implies the possibility of lateral motion in front of the vehicle. Lastly,  event risk assessment helps estimate the potential danger of pedestrian action.  

Besides the importance of each individual task, our agreement study revealed that for a large subset of the samples, the model does not correctly predict both tasks at the same time, partially due to the absence of necessary cues for accurate prediction of either task. This finding suggests that in such scenarios, different tasks can play a complementary role for understanding pedestrian behavior. 

\subsection{Model performance}
Our data analysis and empirical evaluations of existing pedestrian prediction models showed that 1) performance of the models on different tasks is not similar, 2) the ranking of the models varies across different tasks and metrics, and 3) performance on each task is not necessarily impacted by the same factors. As a result, models trained on different tasks (e.g., intention and action) are not directly comparable. 

The new per-instance metrics revealed the lack of temporal consistency in model predictions, even within the short span of $2s$. In intelligent driving systems, such performance fluctuations can lead to irrational behavior by the vehicle or its driver and, consequently, pose risks to other road users. These consistency issues should be remedied in the future works, perhaps, by enforcing instance-wise temporal continuity during training and minimizing the effect of spurious correlations with contextual elements.

In this work, we primarily focused on the input modality of the prediction models and showed that the models that rely more on visual context tend to be more successful on intention estimation and dynamics-oriented models perform better on action and risk tasks. In the future, this analysis can be extended to evaluating the contributions of different modules and interaction between the tasks in a single framework, such as multitasking \cite{rasouli2021bifold,rasouli2023pedformer} or chain reasoning \cite{rasouli2019pie}.

\subsection{Factors that matter for each task}
Our factor analysis highlighted that various contextual elements affect model performance differently on each task. While action is predominantly influenced by dynamics and environmental factors, performance on intention estimation is mainly influenced by pedestrian state. This outcome suggests that intention estimation is inherently a more challenging task as it is not directly influenced by the surrounding environment of the pedestrians Hence, intention estimation models should also capture more subtle behavioral cues, such as the proximity of the pedestrian to the road, their orientation with respect to the road and the ego-vehicle (e.g. looking at the traffic), their closeness to other objects (e.g. transit station) that may reveal their intention (e.g. taking a ride), their other activities (e.g. talking on the phone or another person), and many more elements that require context analysis and spatial reasoning.

Lastly, we examined the variability of the model performance on a single task w.r.t. different data properties. Although some major trends were observed, according to behavioral literature \cite{Rasouli_2019_ITS}, there are many more factors that potentially impact the motives and behaviors of pedestrians. As a result, a more fine-grained analysis based on multiple contextual factors is needed, but was not possible here due to data limitations. Future data collection efforts could mitigate this issue by ensuring sufficient data scale and diversity.

\bibliographystyle{IEEEtran}
\bibliography{references}

\end{document}